\documentclass[conference, 9pt]{IEEEtran}
 
\IEEEoverridecommandlockouts                              

\usepackage[usenames,dvipsnames]{color}
\usepackage{amsmath,amssymb,amsfonts,wasysym}
\usepackage{soul} 
\usepackage[english]{babel}
\usepackage{amsmath}
\usepackage{amsbsy}
\usepackage{gensymb}
\usepackage{graphicx}
\usepackage{subfigure}
\usepackage{xcolor}
\usepackage{color}
\usepackage{float}
\usepackage[backend=biber,style=numeric-comp,sorting=none]{biblatex}
\usepackage{placeins}
\input{macro_marchand_survey.sty}

\title{\LARGE \bf Visual Servoing from Deep Neural Networks}
\author{
Quentin Bateux\textsuperscript{1}, Eric Marchand\textsuperscript{1}, J\"urgen Leitner\textsuperscript{2}, Fran\c cois Chaumette\textsuperscript{3}, Peter Corke\textsuperscript{2} \thanks{
\textsuperscript{1} Quentin Bateux and Eric Marchand are with Universit\'e de Rennes~1, IRISA, Inria, Rennes, France,
{\tt quentin.bateux@irisa.fr, eric.marchand@irisa.fr} \newline
\textsuperscript{2} J\"urgen Leitner and Peter Corke are with the Australian Centre for Robotic
Vision (ACRV), Queensland University of Technology (QUT), Brisbane,
Australia,
{\tt j.leitner@roboticvision.org, peter.corke@qut.edu.au} \newline
\textsuperscript{3} Fran\c cois Chaumette is with Inria, IRISA, Rennes, France, {\tt Francois.Chaumette@inria.fr}
}
}
\date{}

\begin{document}

\maketitle
\thispagestyle{empty}
\pagestyle{empty}
\newcommand{\xp}{\textit{\textbf{x}}_{p}}
\newcommand{\dxpdr}{\dfrac{\partial\xp}{\partial\textbf{r}}}
\newcommand{\md}[1]{\textcolor{blue}{#1}} 
\newcommand{\tmd}[1]{\textcolor{red}{\st{#1}}} 
\renewcommand{\tmd}[1]{} 

\newcommand{\pos}{{\bf r}}
\newcommand{\vitesse}{{\bf v}}
\newcommand{\pIStar}{\boldsymbol{p_{I^*}}(i)}
\newcommand{\pI}{\boldsymbol{p_I}(i)}
\newcommand{\IBarre}{{\bf I}}
\newcommand{\histoX}[1]{\phi( #1 -i)}
\newcommand{\Lx}{L_{\xv}}
\newcommand{\dI}{\nabla I}
\newcommand{\dIx}{\nabla_x {\bf \IBarre}}
\newcommand{\dIy}{\nabla_y {\bf \IBarre}}
\newcommand{\ddr}[1]{\frac{\partial}{\partial \boldsymbol{r}}\left( #1 \right)}
\newcommand{\ddi}[1]{\frac{\partial}{\partial i} \left( #1 \right)}
\newcommand{\sumBin}[1]{\sum_i^{N_c} \left( #1 \right)}
\newcommand{\sumPix}[1]{\sum_{\xv}^{N_{\xv}} \left( #1 \right)}
\newcommand{\xv}{\text{\verb |x| }}  
\renewcommand{\xv}{\text{x}}

\def\baselinestretch{0.97}

\begin{abstract}
We present a deep neural network-based method to perform high-precision, robust and real-time 6~DOF visual servoing.
The paper describes how to create a dataset simulating various perturbations (occlusions and lighting conditions)
from a single real-world image of the scene.
A convolutional neural network is fine-tuned using this dataset to estimate the relative pose between two images of the same scene.
The output of the network is then employed in a visual servoing control scheme.
The method converges robustly even in difficult real-world settings with strong lighting variations and occlusions.
~A positioning error of less than one millimeter is obtained in experiments with a 6~DOF robot.

\end{abstract}

\section{Introduction}
Visual perception is important for humans and robots alike, it provides rich and detailed information about the environment the agent is moving in. 
The goal of visual servoing techniques is to control a dynamic system, such as a robot, by using the information provided by one or multiple cameras~\cite{Hutchinson96a,Chaumette06a}. Classical approaches to visual servoing rely on the extraction, tracking and matching of a set of visual features. These features, generally points, lines, or moments, are  used as inputs to a control law that positions (or navigates) the robot in a desired pose.
Many control strategies
have been proposed over the years, in particular neural networks have been considered when designing control schemes early on~\cite{Krose90,Wells96a}.
 
The tracking and matching of such features, especially given the rich and detailed information stemming from cameras, is a difficult task.
While there has been progress in extracting the relevant features, a technique called direct visual servoing was introduced recently
for exploiting the full image, requiring no feature extraction~\cite{Collewet11a}.
The main drawback of this direct approach is its small convergence domain compared to classical techniques.
This is due to high non-linearities between the image information and the 3D motion.
To remedy this issue we herein propose the use of a trained deep neural network to perform the extraction of features
and estimation of the current image's pose relative to the desired.

More precisely, the following contributions are described herein:
\begin{itemize}
\item re-purposing a commonly used deep neural network architecture, pre-trained for object classification, to perform relative camera pose estimation
\item a novel training process, based on a single image (acquired at a reference pose), which includes the fast creation of a dataset using a simulator allowing for quick fine-tuning of the network for the considered scene. 
It also enables simulation of lighting variations and occlusions in order to ensure robustness.
\item integrating the network with a position-based visual servoing control scheme  robust to occlusions and variations in the lighting
\item achieving precise positioning (sub-mm accuracy) on a 6~DOF robotic setup on planar scenes.
\end{itemize}

\begin{figure}
\centering
\includegraphics[width=.99\columnwidth]{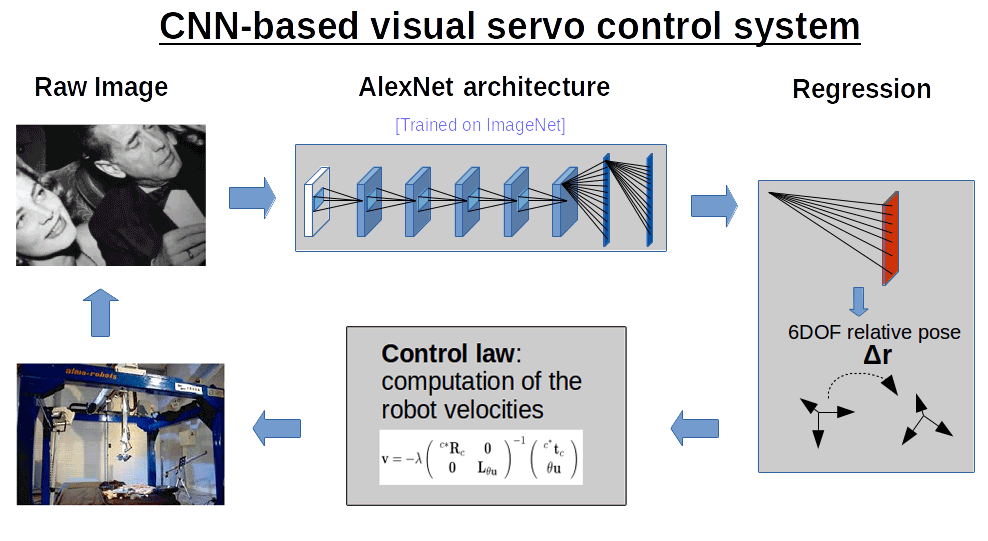}
\caption{Overview of the proposed CNN-based visual servo control system}
\label{alexnet-visabs}
\end{figure}

\section{Related Work}
Visual Servoing (VS) techniques have been applied to a variety of robotic tasks, including reaching, docking and navigation. 
While most of these approaches require a feature extractor or model tracker, direct VS was introduced to make use of the information available in full images~\cite{Collewet11a}. 
The principle is to directly compare the current image with the desired image as a whole, avoiding the classical but error-prone feature detection, tracking and matching steps.
Various control laws have been proposed in order to improve the robustness of direct VS approaches by varying descriptors or the cost functions, such as mutual information~\cite{Dame11b}, histogram distances~\cite{Bateux17a} or mixture of Gaussians~\cite{Caron16a}.

The main drawback of these dense approaches is their small convergence domain,  due to high non-linearities between the image information and the 3D motion. To remedy this issue, combining direct VS with other approaches has been proposed recently, such as the use of particle 
filtering in the control scheme~\cite{Bateux16a} or to consider photometric moments to retain geometric information~\cite{Bakthavatchalam13a}.
In this paper, we propose using deep learning and Convolutional Neural Networks (CNN) to direct VS schemes.
Over the last years deep neural networks, especially CNNs,
have progressed the state-of-the-art in a number of computer vision tasks: 
image classification for object recognition~\cite{Krizhevsky12a}, inferring a depth map from a single RGB image~\cite{Eigen14a}, computing displacement through homography estimation~\cite{Detone16a}, and performing camera relocalization~\cite{Kendall15a}. 
Deep learning has also started to become more prominent in robotics. For example, CNNs have also been trained for predicting grasp locations~\cite{Pinto16a}. 
Recent progress in deep learning reaching tasks has achieved promising results:
learning complex positioning tasks facilitated by vision~\cite{Levine16a,Finn16a}, coupled  with reinforcement learning~\cite{Lange12a}, and purely from vision without the use of any prior knowledge~\cite{Zhang15a}.
A hindrance to wide spread use is that these systems require large datasets and long training times. Our proposed approach partly overcomes this issue.

Due to the availability of easy to use frameworks, it is possible
to construct, train and share deep neural networks and 
to build on networks already trained on very large datasets of millions of images.
It is possible to re-purpose existing networks (with robust global descriptors already embedded in the lower layers) by using fine-tuning techniques (such as in~\cite{Karayev2013a}).
During tuning only the task-specific upper layers of the network are re-trained to perform a different task. 
In our case, rather than performing classification, the network is re-purposed to estimate the relative pose with respect to a desired image.
The main advantage of using a network compared to the previous methods is that the deep learning approach can both create appropriate feature descriptors and also combine them in an optimized way for the designated task. This set of techniques is also well suited to real-time applications since once the training has been performed offline, the online computation of the task is fast (50ms on a middle-end graphics card), with little memory overhead, and most of all constant in term of computation costs, independently of the size and complexity of the dataset it was trained on.

\section{CNN-based Visual Servoing Control Scheme}

The aim of a camera/end-effector positioning task is to reach a desired pose ${\bf r^*}$ starting from an arbitrary initial pose ${\bf r}$ 
(both $\in se(3)$). 
Image-based visual servoing (IBVS) is a method to control camera motion to minimize the positioning error between ${\bf r}$ and ${\bf r^*}$ in the image space~\cite{Chaumette06a}.

\subsection{From visual servoing to direct visual servoing}
\label{DVS}

Considering the actual pose of the camera ${\bf r}$, the
problem can therefore be written as an optimization process:
\begin{equation}
\widehat \pos = \arg\min_\pos \rho(\pos, \pos^*)
\label{classicalasserv}
\end{equation}
where $\widehat \pos$ is the pose reached after the optimization (servoing) process, the closest possible to $\pos^*$ if the system has converged (optimally $ \widehat \pos = \pos^* $), and $\rho(.)$ is arbitrary cost function with a global minimum. For example, considering a set of geometrical
features ${\bf s}$ extracted from the image, the goal is to minimize the error between ${\bf s(r)}$ and the desired configuration ${\bf s^*}$, which leads, by using as cost function the Euclidean norm of the difference between $\bf s$ and $\bf s^*$, to:
\begin{equation}
\widehat \pos = \arg\min_\pos \; \| \bf s(\pos) - s^* \|_2
\label{classicalasservssstrar}
\end{equation}
Visual servoing is classically achieved by iteratively applying a velocity command to the robot.
This usually requires the knowledge of the interaction matrix  
${\bf L_s}$ that links the temporal variation of visual features $\bf \dot{{\bf s}}$ to the camera velocity $\vitesse$: 
\begin{eqnarray}
\dot{\bf s}(\pos) = {\bf L_s}\vitesse.
\end{eqnarray}
This equation leads to the expression of the velocity that needs to be applied to the robot. The control law is classically given by~\cite{Chaumette06a}:
\begin{equation}
 {\bf v} = - \lambda {\bf L^+_{\bf s}}\bf (s(\pos) - s^*)
\end{equation}
where $\lambda$ is a positive scalar and ${\bf L^+_{\bf s}}$ is the pseudo-inverse of the interaction matrix.
\newcommand{\Ib}{{\bf I(r)}}
To avoid the classical but error-prone extraction and tracking of geometrical features (such as points, lines, etc.),
the notion of direct (or photometric) visual servoing has been introduced. 
It considers the image as a whole as the visual feature~\cite{Collewet11a}, ie.\ the set of features {\bf s} becomes the image itself, ${\bf s(r)} = \Ib$. 
The optimization process can then be expressed as: 
\begin{equation}
\widehat \pos = \arg\min_\pos \; \| \bf \Ib - I^* \|_2
\label{opti_luminance}
\end{equation}
where $\Ib$ and $\bf I^*$ are respectively the image seen at the pose $\pos$ 
and the reference image (both composed of $N$ pixels). 

The main issue when dealing with direct visual servoing  is that the 
interaction matrix ${\bf L}_{\bf I}$ is ill-suited for optimization, 
mainly due to the heavily non-linear nature of the cost function,   
 resulting in a small convergence domain. This is the same for the other features that have been considered in direct visual servoing (histogram distances, mutual information, etc.).

\subsection{From direct visual servoing to CNN-based visual servoing}

In this paper, we propose to replace the classical direct visual servoing~\cite{Collewet11a}, as described above, by a new scheme based on a convolutional neural network (CNN). The  network is trained to estimate the relative pose between the current and reference image. Given an image input ${\bf I}({\bf r})$ and the reference image ${\bf I}_0$, let the output of the network be:
\begin{equation}
   {\bf \Delta ^0r} = net_{{\bf I}_0}({\bf I}({\bf r})) 
\end{equation}
with ${\bf \Delta ^0r} = (\T{c_0}{c}, \theta{\bf u})$ the vector representation of the homogeneous matrix $\M{c_0}{c}$ that expresses the current camera frame with respect to the camera frame associated to the reference image. (\textit{Note:} $\theta{\bf u}$ is the angle/axis representation of the rotation matrix $\R{c_0}{c}$.) 

If one wants to reach a pose related to a desired image ${\bf I}^*$, the CNN is first used to estimate the relative pose $\M{c_0}{c^*}$ (from $net_{{\bf I}_0}({\bf I^*})$), and then  $\M{c_0}{c}$ (from $net_{{\bf I}_0}({\bf I})$), from which  ${\bf \Delta ^*r}$ using $\M{c*}{c} = ^{c_0}{\bf T}_{c^*}^{-1} \; \M{c_0}{c}$ is obtained.

Using the cost function $\rho(.)$ such as the Euclidean norm of the pose vector in Eq.~\eq{classicalasserv}, the minimization problem becomes 
\begin{equation}
\widehat \pos = \arg\min_\pos \; \| {\bf \Delta ^*r} \|_2
\label{opti_luminance_new}
\end{equation}
which is known to present excellent properties~\cite{Chaumette06a}.
Indeed, the corresponding control scheme belongs to pose-based visual servoing, which is globally asymptotically stable (ie.\ the system converges whatever the initial and desired poses are), provided the estimated displacement ${\bf \Delta ^*r}$ is stable and correct enough. We recall that IBVS, and thus the schemes based on Eq.~\eq{classicalasservssstrar} and~\eq{opti_luminance_new}, can only be demonstrated as locally asymptotically stable for 6~DOF (ie.\ the system converges only if the initial pose lies in a close neighborhood of the desired pose). With our approach, the stability and convergence issues are thus moved from the control part to the displacement estimation part.

From ${\bf \Delta ^*r}$ provided by the CNN, it is immediate to compute the camera velocity using a classical control law~\cite{Chaumette06a} :
\begin{equation}
  {\bf v} = - \lambda  \left(\begin{array}{c}  \R{c}{c^*} \; \T{c^*}{c} \\\theta{\bf u} \end{array}\right) 
\end{equation}
By computing this velocity command at each iteration, it is then possible to servo the robot toward a desired pose solely from visual inputs.

\subsection{Designing and training a CNN for visual servoing}

In order to keep training time and the dataset size low, we present a method using a pre-trained network.
Pre-training is a very efficient and widespread way of building on CNNs trained for a specific task. If a new task is similar enough, fine-tuning can be performed on the CNN so it may be employed in a different task. 
Since we work on natural images in a real-world robotic experiment, a pre-trained AlexNet~\cite{Krizhevsky12a} was chosen as a starting point. This network was trained on 1.2 million images from the ImageNet set, with the goal of performing object classification (1000 classes). 
While we are not interested in image classification, works such as~\cite{Karayev2013a} showed that it is possible to re-purpose a network by using the learned image descriptors embedded in the lower layers of an already trained AlexNet.
This process, commonly referred to as fine-tuning, is based on the idea that certain parts of the network are useful to the new task and therefore can be transferred.
Particularly the the lower layers (basic image feature extractors) will perform similar functionality in our relative pose estimation task.
Only the upper layers require adaptation. Fine-tuning reduces training time (and data requirements).

We substitute the last layer -- originally outputting 1000 floats with the highest representing the correct class -- by a new layer that output 6 floats, ie.\ the 6~DOF pose.
By replacing this layer, learned weights and connections are discarded and the new links are initialized randomly (see Figure~\ref{alexnet-visabs}).
The resulting net is trained by presenting examples of the form (${\bf I}$, ${\bf r}$), where ${\bf I}$ is a raw image, and ${\bf r}$ the corresponding relative pose as a training label.
Since our training deals with distance between two pose vectors, we choose Euclidean cost function for network training, replacing the commonly used soft-max cost layer for classification, of the following form:
\begin{equation}
loss({\bf I}) = ||\widehat{\T{c_0}{c}} - \T{c_0}{c}||_2 + \beta ||\widehat{\theta{\bf u}} - \theta{\bf u}||_2
\end{equation}
$\widehat{\T{c_0}{c}}$ and $\widehat{\theta{\bf u}}$ respectively are the estimation of the translation and rotation displacements relatively to the reference pose. $\beta = 0.01$ is a scale factor to harmonize the amplitude of the translation (in m) and rotation (in degrees) displacements to facilitates the learning process convergence.

Starting from the trained AlexNet available for use with the Caffe library~\cite{Jia14a}, the network was then fine-tuned. For this a new scene specific dataset of 11000 images with a variety of perturbations is created (as described in the next section).
Using Caffe the network was trained with a batch size of 50 images over 50 training epochs. 

The proposed method can be used with any kind of CNN network trained on images,
therefore taking advantage of future developments in deep learned image classification. A thorough comparison of architectures is left for future work.

\section{Designing a Training Dataset}
The design of the training dataset is the most critical step in the process of training a CNN, as it affects the ability of the training process to converge, as well as the precision and robustness of the end performances. 
As stated above we propose to fine-tune of a pre-trained network. 
Gathering real-world data is often cumbersome and sometimes unsuitable depending of the environment where the robot is expected to operate in. Furthermore, it can be difficult to re-create all possible conditions within the real-world environment.
In this section we describe how simulated data allows us to generate a virtually unlimited amount of data. In addition we show how a variety of perturbations can be added which leads to satisfactory results without lengthy real-world data acquisition.

\begin{figure}[tbp]
\centering
\includegraphics[width=0.99\columnwidth]{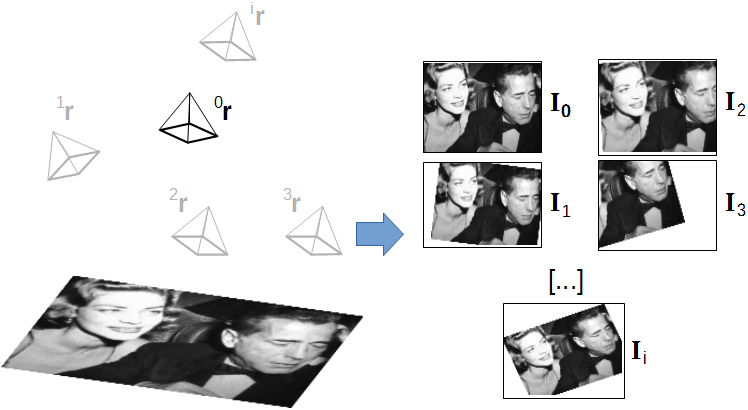}
\caption{3D plane with the projected reference image and multiple virtual cameras to generate multiple views of the scene.}
\label{simulator}
\end{figure}

\begin{figure*}[tb!]
\centering
\includegraphics[width=2\columnwidth]{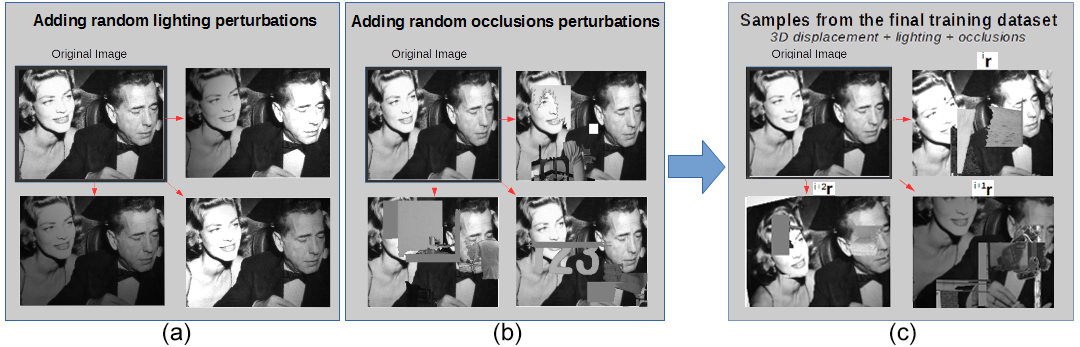}
    \caption{Overall process to create a training set from the original input image: (a) examples after applying local illumination changes; (b) examples after adding super-pixels from random images as occlusions; (c) examples from the final dataset after applying all perturbations.}
    \label{perturbations}
\end{figure*}

\subsection{Creating the nominal dataset}

The nominal dataset is the base of the training dataset. It contains all the necessary information for the CNN to learn how to regress from an image input to a 6~DOF relative pose. 
We will be adding various perturbations later on to ensure robustness 
when confronted with real-world data.

In our design, the nominal dataset is generated from a single real-world image ${\bf I_0}$ of the scene. This is possible by relying on simulation, in order to create images as viewed from virtual cameras. 
Figure~\ref{simulator} illustrate the image projected on a 3D plane and the varying (virtual) camera poses.
This procedure generates datasets of thousand of images quickly (less than half an hour for 11k images)
eliminating the time-consuming step of gathering real-world data.
In comparison, 700 robot hours were necessary to gather 50k data points for a single task in~\cite{Pinto16a}. 

The procedure to create the synthetic training dataset is then as follows (see also Figure~\ref{perturbations}):
\begin{itemize}
\item acquire a single image ${\bf I_0}$ at pose ${\bf ^{0}r}$, in order to get the camera characteristics and scene appearance 
\item create a 10,000 elements dataset, consisting of tuples ($^i{\bf r}, {\bf I}_i$), through virtual camera views in simulation (as illustrated in Figure~\ref{simulator}).
The first 10,000 virtual camera poses are obtained using a Gaussian draw around the reference pose $^0{\bf r}$, in order to have an appropriate sampling of the parameters space (the 6~DOF pose). The scene in the simulator is set up so that the camera-plane depth at $^{0}{\bf r}$ is equivalent to 20cm, and the variances for the 6DOF Gaussian draw are such as (1cm, 1cm, 1cm, 10\degree, 10\degree, 20\degree), respectively for the ($\text{t}_x, \text{t}_y, \text{t}_z, \text{r}_x, \text{r}_y, \text{r}_z$) DOF.

\item the dataset is appended with 1,000 more elements. These are created by a second Gaussian draw with smaller variances (1/100 of the first draw). The finer sampling around ${\bf ^0r}$ enables the sub-millimeter precision at the end of the robot motion.
\end{itemize}

\subsection{Adding perturbations to the dataset images}
\label{sec:perturbations}
\label{sec:illumination_variation}
\label{sec:occlusions}

In order to obtain a more robust process, two main perturbations were modeled and integrated in the dataset, namely, lighting changes (both global and local) and occlusions. We assume the scene to be static under nominal conditions for each experiment (ie.\ no deformations or temporal changes in the structure).

\subsubsection{Modeling illumination variations with 2D Gaussian functions}

Lighting conditions are a common problem when dealing with real-world images. 
These are of global and local nature. 
In order to model the global light, one can simply alter the pixel intensities by considering affine variation of the intensities.
Local lighting changes are more challenging and to obtain realistic synthetic images
time-consuming rendering algorithms are required.
We alleviate this issue by working with planes in 3D space only, allowing to model lights as local 2D light sources and get realistic results.
For each image chosen to be altered, the following mixture of 2D Gaussians is applied at each pixel $(x, y)$:
  \begin{equation}
    {\bf I}_{{l}}(x, y) = \sum_{l=1}^{N_{lights}}{\bf I}(x,y) f_l(x,y)
  \end{equation}
Each 2D Gaussian in turn can be modelled as 
  \begin{equation}
    \label{eq:2dgaussian}
    f_l(x,y)=Ae^{-\left( \frac{(x-x_0)}{2\sigma^2_x} + \frac{(y-y_0)}{2\sigma^2_y} \right)}
  \end{equation}
where  $(x_0, y_0)$  (in pixel units) corresponds to the projection of the center of the simulated directional light source,  gain $A$  to its intensity, and  $(\sigma_x, \sigma_y)$  to the spread along each axis. An example of the resulting images can be seen in Figure~\ref{perturbations}(a). 
We purposely let out the modeling of specularities, as the material and reflection properties are unknown.
Our method will handle them as a sub-class of occlusions (see next section).

\begin{figure}[tb]
  \centering
    \includegraphics[width=\columnwidth]{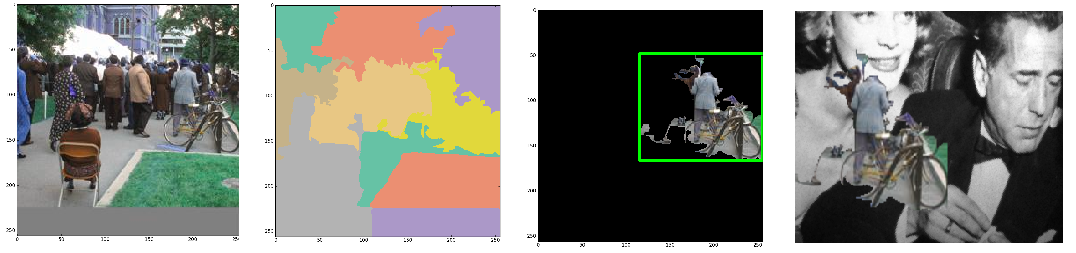}
  \caption{Synthetic occlusion generation: on an arbitrary images in the {\it{Label-Me}} dataset (left) segmentation is performed. A segmented cluster is selected at random. It provides a coherent ``occlusion'' patch, which is merged with a dataset image and added to the dataset (last image). }
  \label{superpixel}
\end{figure}

\subsubsection{Modeling occlusions by superimposing coherent pixel clusters from other image datasets}
Dealing with occlusions is challenging due to the potential variety in size, shape and appearance that can appear in a real environment. Additionally, when training a CNN, one has to be careful to create the training set with a variety of perturbations included. This is to prevent the network from over-fitting on the presented examples and thus being unable to handle real world occlusions.
We present here a procedure to provide the network with a realistic set of occlusions with an adequate range in size, shape and appearance.

To address this issue, we are adding clusters of pixels -- representing a coherent part of an image -- from other datasets and superimpose them on the previously generated images.
To create somewhat realistic conditions real world images were preferred over synthetic occlusion images.
These images provide a variety of scenes that represent interesting and varied occlusions, rather than those generated from geometrical or statistical methods. 
Herein the {\it{Label-Me}} dataset~\cite{Russel08a} containing thousands of outdoor scene images was chosen.
The scenes contain a variety of objects in a wide range of lighting conditions.
We then applied the following work-flow (illustrated in Figure~\ref{superpixel}) to each of the images in our simulated dataset that we want to alter:
\begin{itemize}
\item select randomly one image from the {\it{Label-Me}} dataset;
\item perform a rough segmentation of the image by applying the SLIC super-pixel \cite{Achanta12} segmentation algorithm creating coherent pixel groups (implementation available in OpenCV) 

\item we select a random cluster from the previous step, and then insert this cluster into the image to alter at a random position.
\end{itemize}

This method allows us to get a training dataset with randomly varied occlusions such as illustrated in Figure~\ref{perturbations}(b).
By stacking the two described perturbations on our initial nominal dataset, we are able to generate a final training dataset with all the desired characteristics, as shown in Figure~\ref{perturbations}(c).

\section{Experimental Results on a 6~DOF Robot}
This section describes a set of experiments performed on an Afma 6~DOF gantry robot in a typical eye-in-hand configuration. At the beginning of each experiment, the robot is moved to an arbitrary starting pose $\bf{r_0 }$ and the task is to navigate the robot back to a position defined by a desired image.

\subsection{Nominal Conditions}
\begin{figure*}[t]
  \centering
    \subfigure{
    \label{nominalA}
    \includegraphics[width=0.4\columnwidth]{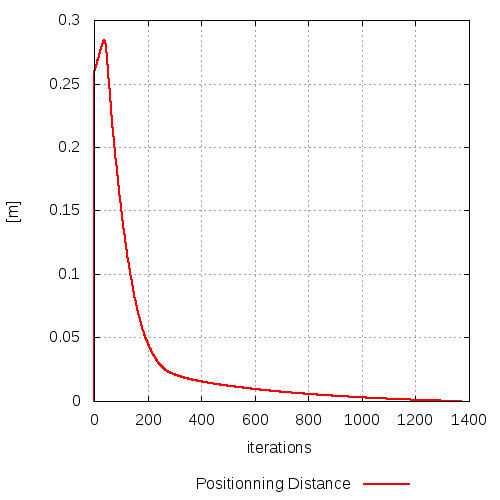}a
  }
  \subfigure{
    \label{nominalB}
    \includegraphics[width=0.4\columnwidth]{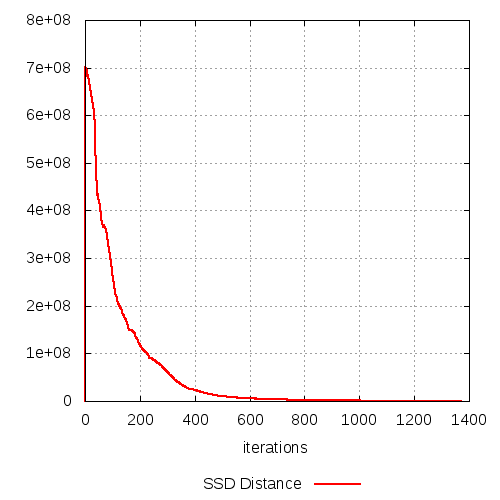}b
  } 
  \subfigure{
    \label{nominalC}
    \includegraphics[width=0.4\columnwidth]{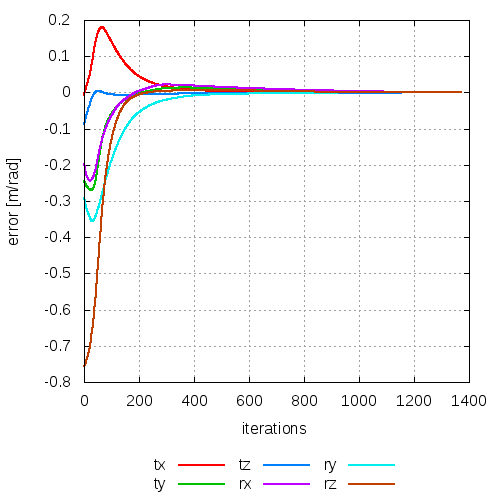}c
  } 
  \subfigure{
    \label{nominalE}
    \includegraphics[width=0.4\columnwidth]{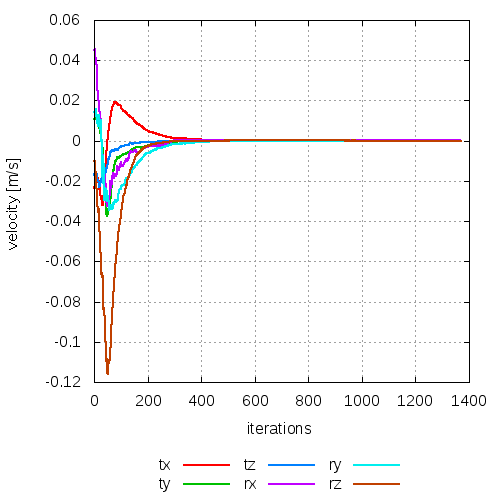}d
  }

  \subfigure{
    \label{nominalF}
    \includegraphics[width=0.4\columnwidth]{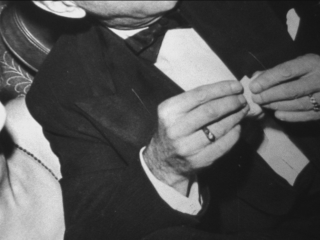}e
  } 
  \subfigure{
    \label{nominalG}
    \includegraphics[width=0.4\columnwidth]{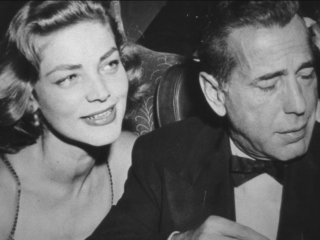}f
  } 
  \subfigure{
    \label{nominalH}
    \includegraphics[width=0.4\columnwidth]{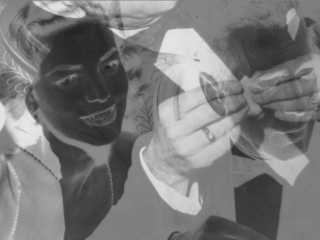}g
  }
  \subfigure{
    \label{nominalI}
    \includegraphics[width=0.4\columnwidth]{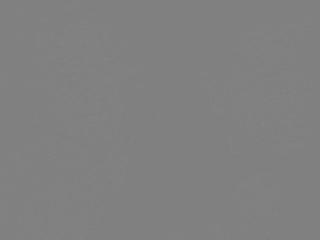}h
  }
  \caption{CNN-based visual servoing on a planar scene; (a) Positioning error; (b) SSD distance; (c) Translational and rotational errors; (d) Camera velocities in m/s and rad/s; 
  (e) Image at initial pose $\bf{I_0}$; (f) Image at final pose ${\bf I^( \widehat \pos)}$; (g) Image error ${\bf I_0 - I^*}$ at initial pose; (h) Image error ${\bf I^( \widehat \pos) - I^*}$ at the final pose}
  \label{CNN_Finer}
\end{figure*}

In terms of pose offset, the robot has to perform a displacement given by ${\bf \Delta ^0r} = (\T{c_0}{c}, \theta{\bf u})$:
\begin{align*}
    \T{c_0}{c} = (1 \text{cm}, -24\text{cm}, -9\text{cm}),\\
    \theta{\bf u} = (-10\degree, -16\degree, -43\degree), ~~
\end{align*}
with a distance between the camera and the planar scene of 80cm at the desired pose $\bf{r^*}$. Figure~\ref{nominalG} shows the image at the final pose. Figure~\ref{nominalI} shows the image error between the final and desired image. 
The training of the network with the 11000 images was performed offline. 
Figures~\ref{nominalA} through~{\ref{nominalE} show that our CNN-based direct visual servoing approach converges without any noisy nor oscillatory behaviours when performed in a real-world robotic setting. Furthermore, the position of the system at the end of the motion is less than one millimeter from the desired one.
No particular efforts were made to have ``perfect'' lighting conditions, but also no external lighting variations or occlusions were added. These were introduced in the next experiment.

\subsection{Dealing with perturbations: Light changes and occlusions}
\label{sec:exp_light}
Given the same initial conditions as above additional light sources and external occlusions were added to test the robustness of our approach.
The robot captures a single image at the initial pose, the network is trained again and then our CNN-based direct visual servoing is performed.
While the robot is servoing the light coming from 3 lamps is changed independently, resulting in global and local light changes.
In addition during the experiment various objects are added, moved and removed from the scene in order to create occlusions.

\begin{figure*}[t]
  \centering
    \subfigure{
    \label{perturbedA}
    \includegraphics[width=0.4\columnwidth]{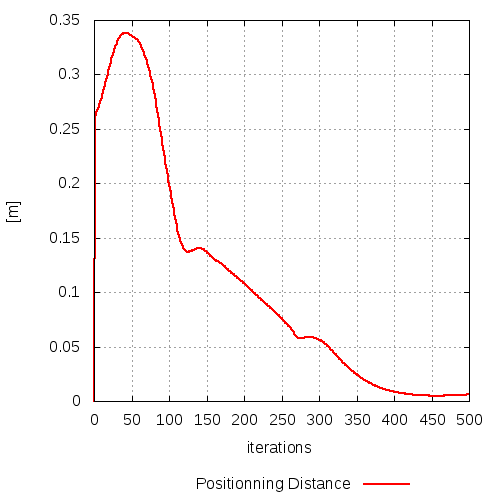}a
  }
  \subfigure{
    \label{perturbedB}
    \includegraphics[width=0.4\columnwidth]{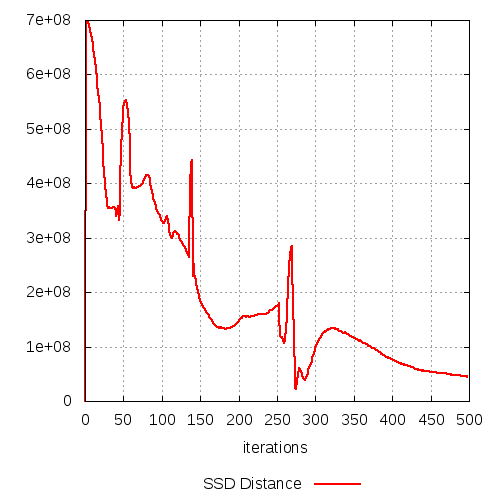}b
  } 
  \subfigure{
    \label{perturbedC}
    \includegraphics[width=0.4\columnwidth]{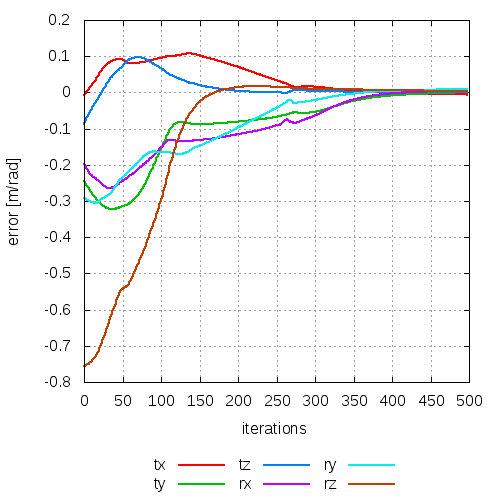}c
  } 
  \subfigure{
    \label{perturbedE}
    \includegraphics[width=0.4\columnwidth]{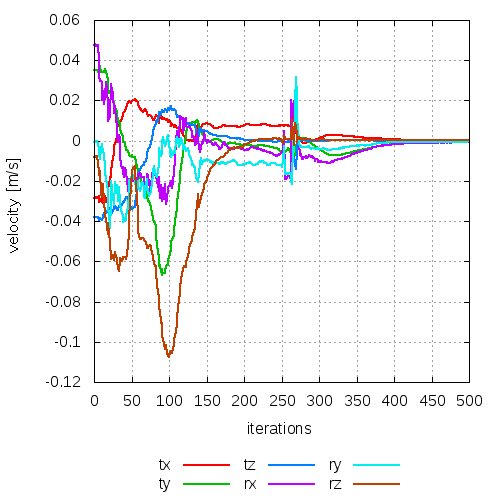}d
  }

  \subfigure{
    \label{perturbedF}
    \includegraphics[width=0.4\columnwidth]{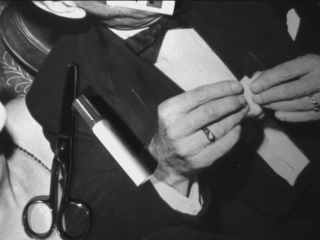}e
  } 
  \subfigure{
    \label{perturbedG}
    \includegraphics[width=0.4\columnwidth]{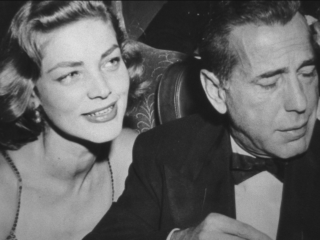}f
  } 
  \subfigure{
    \label{perturbedH}
    \includegraphics[width=0.4\columnwidth]{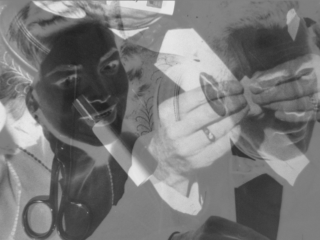}g
  }
  \subfigure{
    \label{perturbedI}
    \includegraphics[width=0.4\columnwidth]{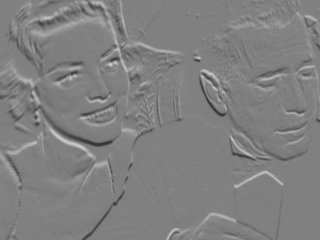}h
  }
  \caption{CNN-based visual servoing on a planar scene with various perturbations. (a) Positioning error; (b) SSD distance; (c) Translational and rotational errors; (d) Camera velocities in m/s and rad/s;
  (e) Image at initial pose $\bf{I_0}$; (f) Image at final pose ${\bf I^( \widehat \pos)}$; (g) Image error ${\bf I_0 - I^*}$ at initial pose; (h) Image error ${\bf I^( \widehat \pos) - I^*}$ at the final pose}
  \label{CNN_Finer_Perturbed}
\end{figure*}

\begin{figure}[tb]
  \centering
  \subfigure{
    \label{videoA}
    \includegraphics[width=0.3\columnwidth]{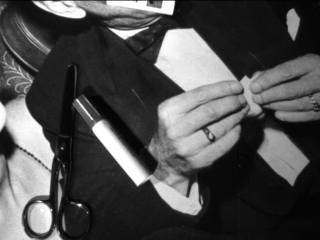}a
  }
  \subfigure{
    \label{videoB}
    \includegraphics[width=0.3\columnwidth]{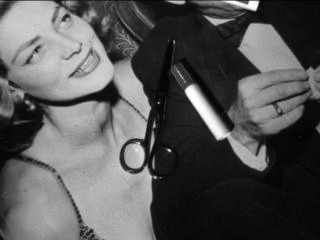}b
  }
  \subfigure{
    \label{videoC}
    \includegraphics[width=0.3\columnwidth]{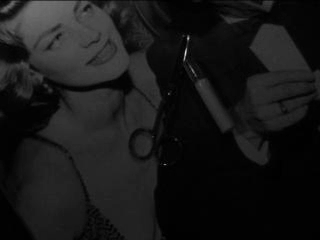}c
  } 
  \subfigure{
    \label{videoD}
    \includegraphics[width=0.3\columnwidth]{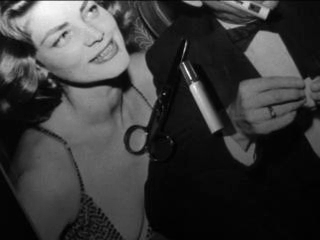}d
  } 
  \caption{Images collected during real-world experiments on our 6DOF robot. Significant occlusions and variation in the lighting conditions can be seen.}
  \label{videoFrames}
\end{figure}

Figure~\ref{CNN_Finer_Perturbed} shows the graphs plotted for this second experiment.
We can see that despite the variety and severity of the applied perturbations, the control scheme does not diverge.
The method instead exhibits only a loss in final precision, ranging from 10cm (in accumulated translation errors) at the worst of the perturbations (samples are depicted in Figure~\ref{videoFrames} and show the various perturbations that occurred) to less than one millimeter error when back in the nominal conditions. A slower convergence is also observed when compared with the nominal conditions experiment.
Additionally, most of the positioning error lies in the coupled translation and rotation degrees-of-freedom tx/ry and ty/rx.
This keeps most of the scene in the camera's field of view by keeping the center of the scene aligned with the optical axis.
The robot can therefore effectively reach the desired pose once the perturbed conditions are removed. 
This resilience is highlighted at iterations 100 and 260 when very strong perturbations occur as the operator's hand briefly occlude most of the scene, inducing a strong spike in both the SSD and the velocities applied to the robot (as the network receives as input only non-relevant information). However, as soon as this perturbation vanishes, the method is able to retrieve instantaneously its converging motion. It is important to note that since no tracking is involved, no elaborate scheme were introduced to deal with sudden loss of information and re-initialization of the method as it is able to regains its performances as soon as the information becomes available again.
It also can be seen that the perturbations observed on the outputs of the network (Figure~\ref{perturbedC}), which are used as inputs of the control scheme, are not synchronous and are less noisy than the perturbations observed in the sum-of-squared-distances (SSD) plot (Figure~\ref{perturbedB}).

\section{Conclusion and Perspectives}

In this paper we presented a new generic method for robust visual servoing from deep neural networks.
We re-purpose pre-trained convolutional neural network (CNN) by substituting the last layer with a new output layer.
Together with a matched general cost function, it enables enable fine-tuning of CNNs for visual servoing tasks. 

Using a regression layer rather than a classification one as output layer re-configures the neural network to estimate the the relative pose to the desired image at each frame.  
Selection of the right dataset is critical for training a neural network, and we herein present an approach to design and collect a synthetic dataset for quick fine-tuning of the network to facilitate visual servoing.
The synthetic data includes multiple views, local illumination changes from simulated 3D light sources, and simulated occlusions using coherent patches from randomly selected real-world image datasets.

We demonstrated the validity and efficiency of this approach with experiments on a 6 DOF gantry robot. The proposed method achieves millimeter accuracy through all 6 degrees of freedom in centimeter- and meter-scale positioning tasks.
Furthermore we have demonstrated that the proposed approach is robust to strong perturbations as lighting variations
~and occlusions
.
The current framework allows a robot to visual servo with respect to a single scene, which forms the basis of the training set.
Changing the application scene only requires synthesis of a new, comparatively small, training dataset and fine-tuning of the network to generate the desired pose estimates.
Future research will focus on extending the proposed method to generalize to multiple scenes (including 3D ones), eventually training a network that provides scene-agnostic relative camera pose estimations.

\printbibliography
\end{document}